\documentclass[letterpaper]{article} 
\usepackage{aaai24}  
\usepackage{times}  
\usepackage{helvet}  
\usepackage{courier}  
\usepackage[hyphens]{url}  
\usepackage{graphicx} 
\usepackage{color}
\usepackage{natbib}  
\usepackage{caption} 
\usepackage{algorithm}
\usepackage{algorithmic}
\usepackage{subfig}
\usepackage{cite}
\usepackage{amsmath}
\usepackage{amssymb}
\usepackage{booktabs}
\usepackage{threeparttable}
\usepackage{multirow}
\usepackage{bm}

\usepackage{newfloat}
\usepackage{listings}
\DeclareCaptionStyle{ruled}{labelfont=normalfont,labelsep=colon,strut=off} 
\floatstyle{ruled}
\newfloat{listing}{tb}{lst}{}
\floatname{listing}{Listing}
\title{Scribble Hides Class: Promoting Scribble-Based Weakly-Supervised Semantic Segmentation with Its Class Label}
\author{
    Xinliang Zhang\textsuperscript{\rm 1,3}\equalcontrib,
    Lei Zhu\textsuperscript{\rm 1-4}\equalcontrib,
    Hangzhou He\textsuperscript{\rm 1-3},
    Lujia Jin\textsuperscript{\rm 1-4},
    Yanye Lu\textsuperscript{\rm 1,3,4}\thanks{Corresponding author, yanye.lu@pku.edu.cn}
}
\affiliations{
    \textsuperscript{\rm 1} Institute of Medical University, Peking University, Beijing, China \\
    \textsuperscript{\rm 2} Department of Biomedical Engineering, Peking University, Beijing, China \\
    \textsuperscript{\rm 3} National Biomedical Imaging Center, Peking University, Beijing, China \\
    \textsuperscript{\rm 4} Institute of Biomedical Engineering, Peking University Shenzhen Graduate School, Beijing, China \\
    zhangxinliang@tju.edu.cn,  \{zhulei, zhuang\}@stu.pku.edu.cn, \{jinlujia, yanye.lu\}@pku.edu.cn

}

\usepackage{bibentry}

\begin{document}

\maketitle

\begin{abstract}
Scribble-based weakly-supervised semantic segmentation using sparse scribble supervision is gaining traction as it reduces annotation costs when compared to fully annotated alternatives. Existing methods primarily generate pseudo-labels by diffusing labeled pixels to unlabeled ones with local cues for supervision. However, this diffusion process fails to exploit global semantics and class-specific cues, which are important for semantic segmentation. In this study, we propose a class-driven scribble promotion network, which utilizes both scribble annotations and pseudo-labels informed by image-level classes and global semantics for supervision. Directly adopting pseudo-labels might misguide the segmentation model, thus we design a localization rectification module to correct foreground representations in the feature space. To further combine the advantages of both supervisions, we also introduce a distance entropy loss for uncertainty reduction, which adapts per-pixel confidence weights according to the reliable region determined by the scribble and pseudo-label's boundary.  Experiments on the ScribbleSup dataset with different qualities of scribble annotations outperform all the previous methods, demonstrating the superiority and robustness of our method. The code is available at \url{https://github.com/Zxl19990529/Class-driven-Scribble-Promotion-Network}.
\end{abstract}

\section{Introduction}

Primarily driven by the availability of extensive pixel-level annotated datasets, the field of semantic segmentation has made remarkable strides in the last decade. However, the challenges of the laborious and time-consuming process of collecting and manually annotating such datasets hinder real-world applications of semantic segmentation. Weakly-supervised semantic segmentation (WSSS) methods utilizing sparse labels have emerged as a prominent trend to overcome this limitation. These methods use annotations at the image, scribble, or bounding box levels as supervision to train the semantic segmentation model. Among them, image-level annotations offer limited spatial supervision, while bounding boxes may lead to overlapping issues when objects are nearby. In comparison, the use of scribble annotations strikes an optimal balance between supervision effectiveness and labor cost~\cite{lin2016scribblesup}. Consequently, scribble-based WSSS has garnered increasing attention in recent years~\cite{liang2022tree,wu2023sparsely}.

\begin{figure}[tbp]
  \centering
  \subfloat[regularization loss]{
      \includegraphics[width=0.46\linewidth]{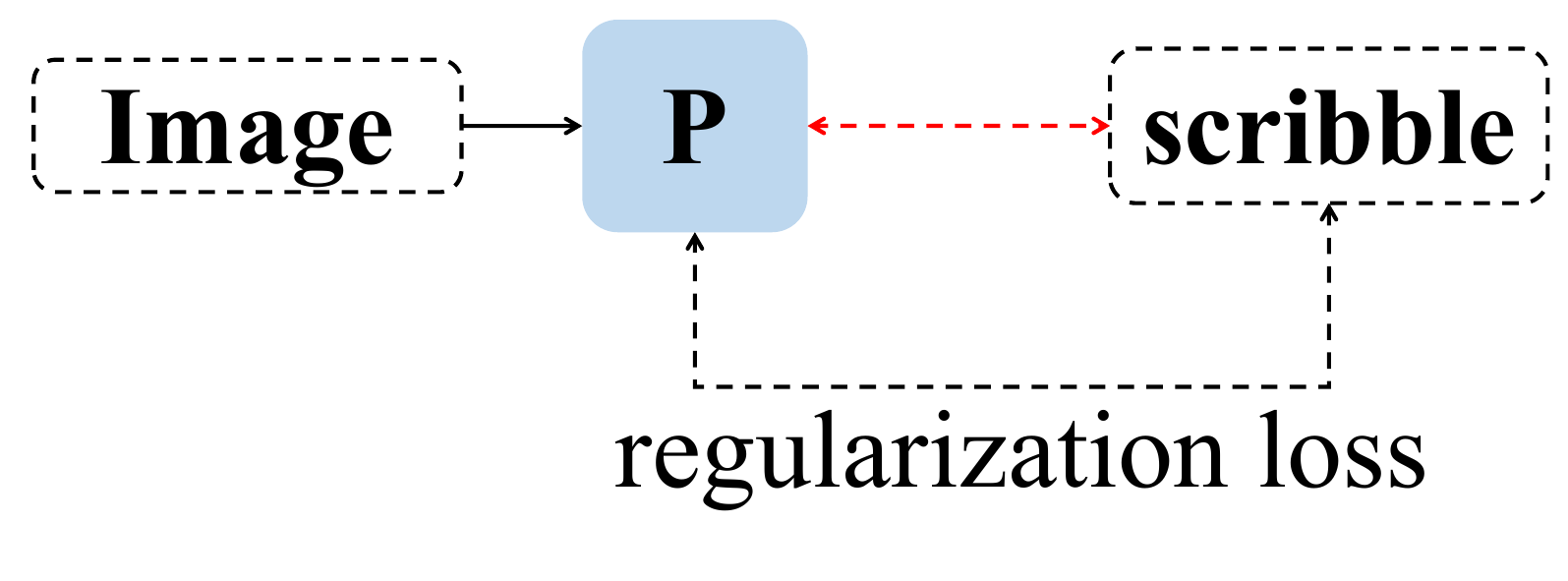}
      }
  \subfloat[consistency learning]{
      \includegraphics[width=0.47\linewidth]{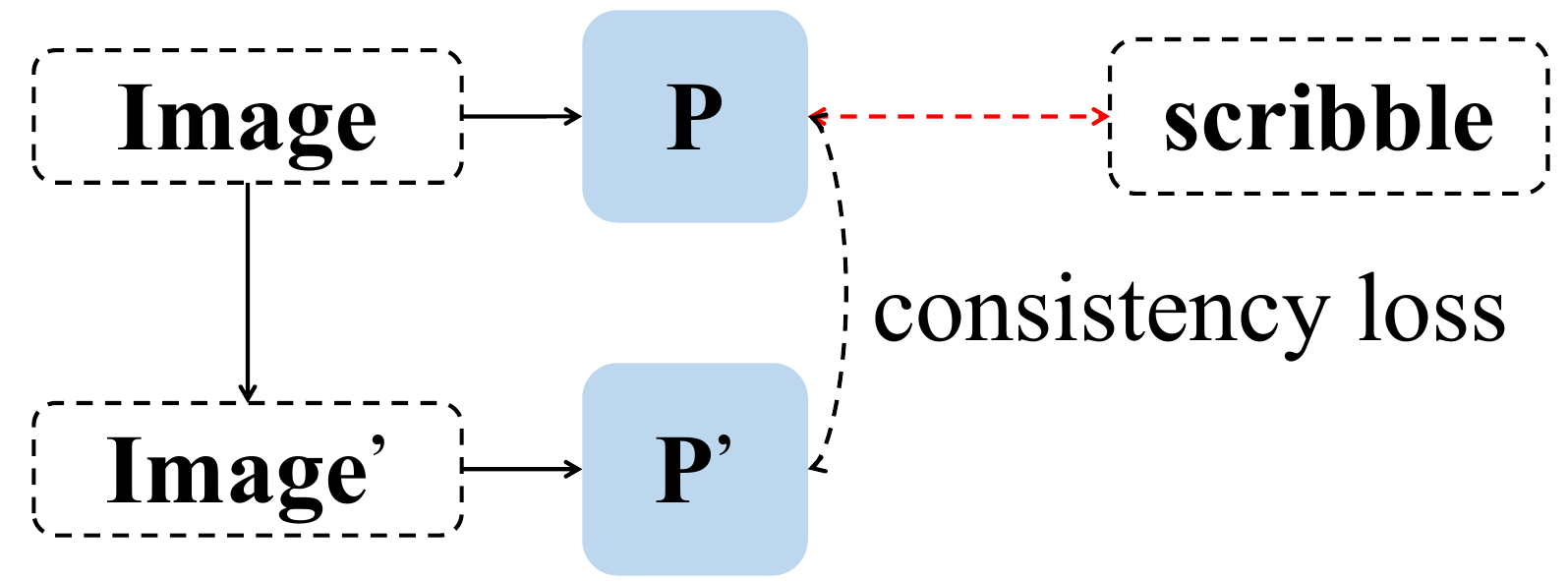}
      }

  \subfloat[label diffusion]{
    \includegraphics[width=0.47\linewidth]{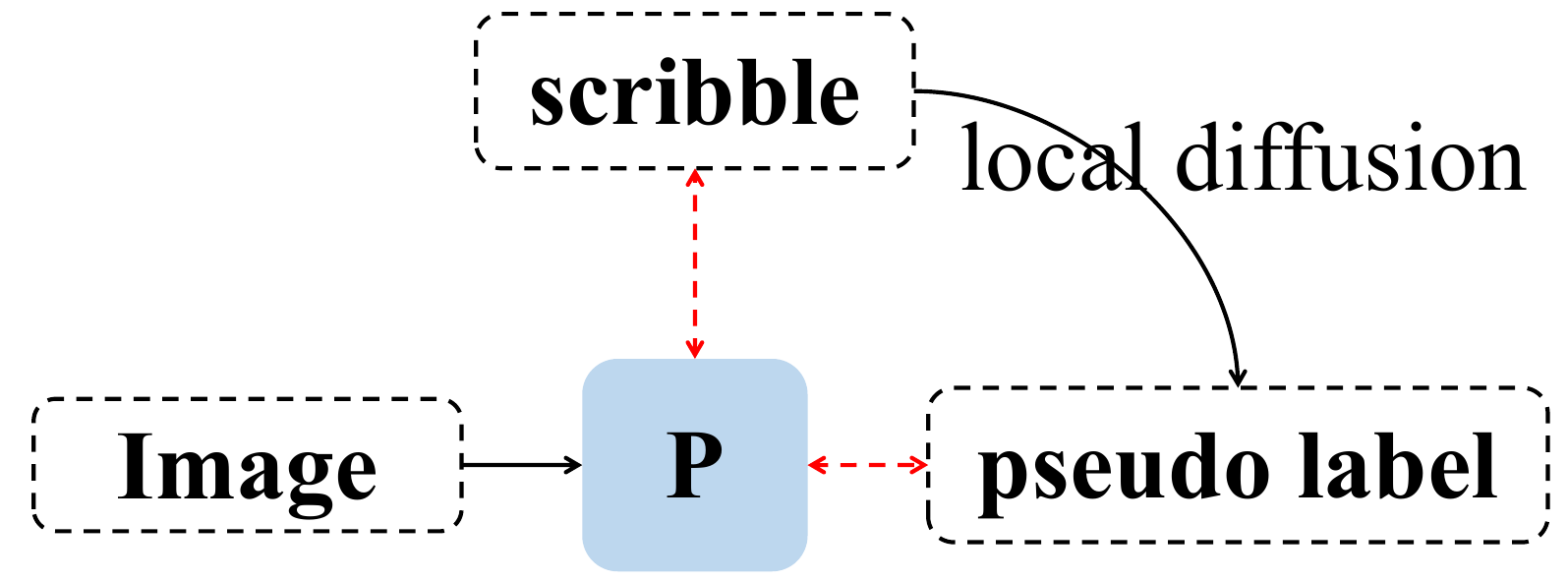}
    \label{fig:ssss_intro_labeldiffusion}
    }
  \subfloat[ours]{
      \includegraphics[width=0.47\linewidth]{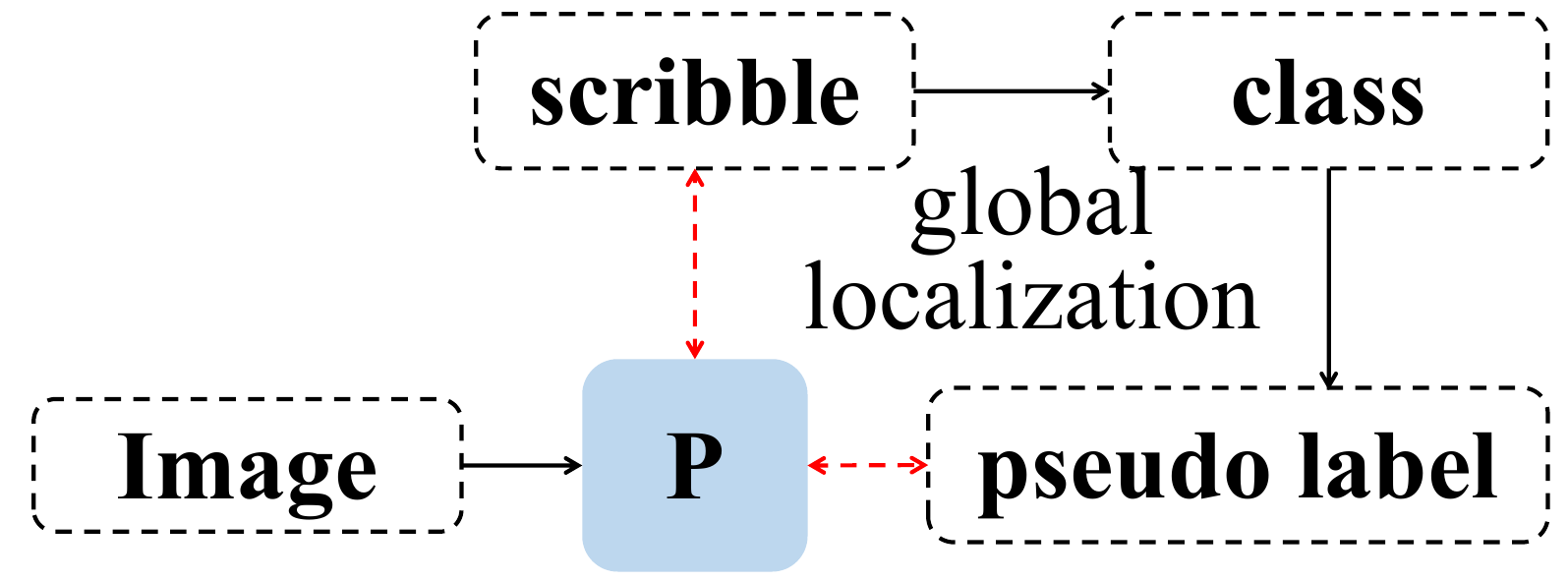}
      }
  \caption{Schematic diagrams of different scribble-based WSSS methods. Existing approaches (a-c) overlooked the class label in scribbles, which provides image-level supervision. ``P" represents the model prediction. The red dashed line represents the supervision relationship.}
  \label{fig:ssss_intro}
\end{figure}

The intrinsic challenge in scribble-based WSSS lies in the partial supervision provided by sparse labels. Existing approaches have attempted to address this issue from three perspectives, namely, regularization loss~\cite{tang2018normalized,tang2018regularized}, consistency learning~\cite{pan2021scribble,wang2022cycle}, and label diffusion~\cite{lin2016scribblesup,wu2023sparsely}, as illustrated in Figure~\ref{fig:ssss_intro}(a-c). Specifically, regularization loss-based methods design specific loss functions to improve the stability of the models. Consistency learning-based approaches aim to capture invariant features to boost fine-grained segmentation performance through consistency loss. However, both methods fail to address the deficiency of pixel-level supervision, leading to limited performance. In contrast, label diffusion-based methods generate pixel-level pseudo-labels by diffusing labeled pixels to unlabeled ones, \emph{i.e.} constructing a graph model on the scribble to generate pseudo-labels for training. However, the diffusion process predominantly relies on local pixel information and fails to exploit the global semantics and class-specific cues of images, which are important for semantic segmentation. In addition, such pseudo-label generation approaches are heavily dependent on the quantity and quality of the scribbles, where the model performance would be undermined when the scribbles are shrunk or dropped as shown in Figrue~\ref{fig:shrink_drop}. In fact, sparse scribbles inherently possess class information, which can offer valuable global semantic clues while enriching scribble-based WSSS supervision. However, this advantage has not been extensively explored in existing scribble-based WSSS researches. 

In light of this, the present paper is dedicated to promoting the performance of scribble-based WSSS with a globally considered pseudo-label. The image-level class labels could be easily obtained from the scribbles, making it feasible to acquire the globally considered pseudo-label via image-level WSSS methods. Previous image-level WSSS methods have demonstrated that image-level class labels prompt models to focus on discriminative areas within an image, which can be used to compensate for the limitations of local cues provided by scribbles. Drawing inspiration from this, we propose a class-driven scribble promotion (CDSP) network for scribble-based semantic segmentation, which utilizes image-level class labels to generate pseudo-labels.

The overview of our method is depicted in Figure~\ref{fig:ssss_intro} (d). We begin by extracting image-level class labels from the scribbles and employing them to train a classification model, subsequently generating the globally considered pseudo-label. We then proceed to train a semantic segmentation model with both scribble and pseudo-label for supervision. By doing so, the inclusion of the image-level class label facilitates the acquisition of global semantic information for pseudo-label generation and further benefits the scribble-based WSSS training. Nevertheless, the noisy supervisions in pseudo-labels may affect the model, where we specifically devise a localization rectification module (LoRM) to address this issue, which corrects foreground representations in the latent feature space by referencing other pixels. To further leverage the advantages of both supervisions, we also introduce a distance entropy loss (DEL) for model uncertainty reduction, where the model prediction is assigned with per-pixel confidence based on the reliable region determined by the scribble and the boundary of the pseudo-label. With these integrated components, our method achieves state-of-the-art (SOTA) performance in scribble-based WSSS. Our contributions can be concluded as:
\begin{itemize}
  \item We present a class-driven scribble promotion network for scribble-based WSSS that utilizes image class information to generate a globally considered pseudo-label. Notably, this is the first approach to exploit image-level class information in the scribble-based WSSS problem.
  \item A localization rectification module is proposed to correct the foreground representations in the latent feature space that are misled by the noisy pseudo-labels. And a distance entropy loss is proposed to excavate the reliable areas based on proximity to scribbles and pseudo-labels.
  \item The proposed method outperforms existing state-of-the-art methods. The extensive experiments on the different qualities of scribbles scribble demonstrate the extraordinary robustness of our method.
\end{itemize}

\begin{figure*}[htbp]
  \centering
  \includegraphics[width=0.85\linewidth]{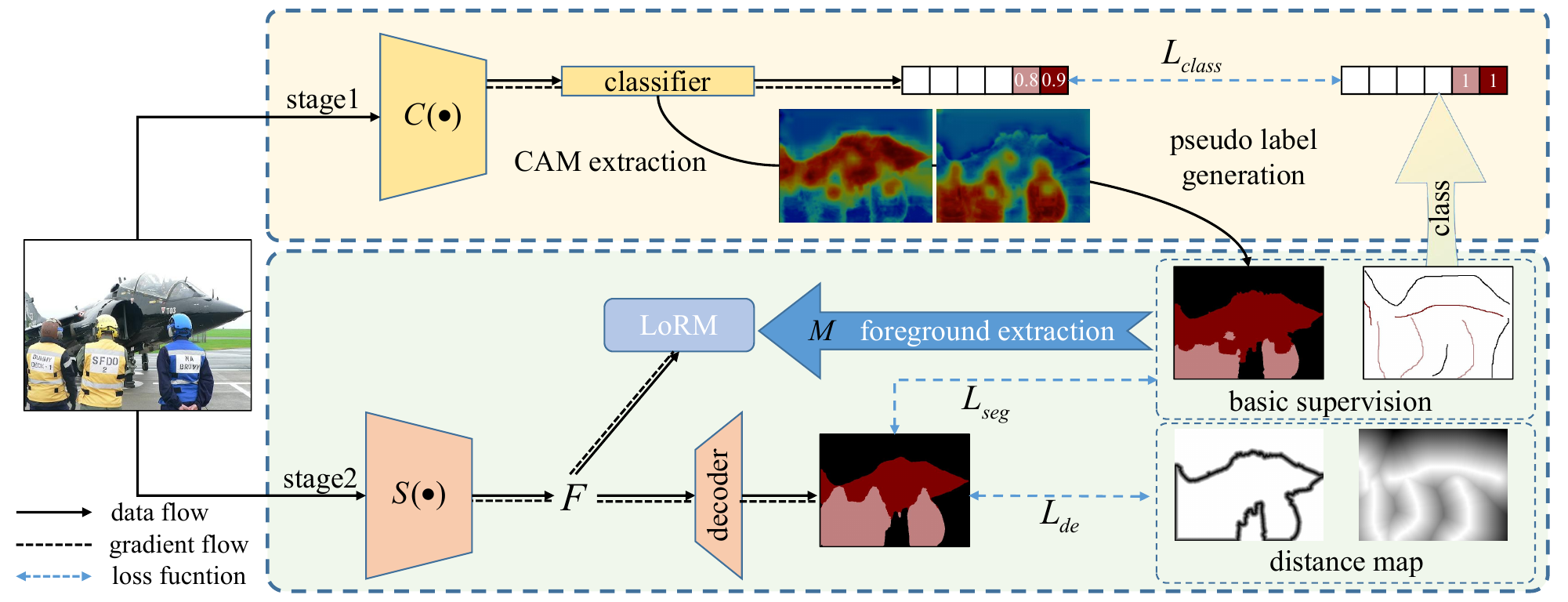}
  \caption{The overview of our method (CDSP). In the first stage, we train a classification model with the image-level class labels extracted from the scribbles to generate the globally considered pseudo-label. Then we train a semantic segmentation model with the globally considered pseudo-label and the scribble label jointly in the second stage. We propose a localization rectification module (LoRM) and a distance entropy loss to assist the training process. }
  \label{fig:shc}
\end{figure*}

\section{Related Works}

\paragraph{Image-level WSSS} The remarkable achievements of early deep learning-based methods in image classification~\cite{simonyan2014very} have spurred numerous works on feature visualization. ~\citet{zhou2016learning} first introduced the class activation map (CAM) technique, which employs global average pooling on deep features to visualize discriminative localization. This technology subsequently catalyzed various efforts to generate semantic pseudo-labels from CAM, facilitating the training of segmentation networks~\cite{kolesnikov2016seed,zhang2021complementary,zhu2023background,zhu2022weakly}. Recently, SEAM~\cite{wang2020self} presented a pixel correlation module that refines current pixel predictions using information on the similar neighbors of the pixel. From another perspective, AFA~\cite{ru2022learning} addressed this problem with transformers leveraging multi-head self-attention for effective long-range modeling. Additionally, ~\cite{ru2023token} developed patch token contrast and class token contrast modules to capture high-level semantics. The intrinsic capability of image-level supervised semantic segmentation to capture global information makes it a promising approach to promote scribble-supervised semantic segmentation.

\paragraph{Scribble-based WSSS} 
Early methods can be traced back to traditional interactive segmentation~\cite{rother2004grabcut,grady2006random},
which employ graphical models to expand the scribble area and extract foreground regions. These methods typically require multiple continuous interactions to extract foreground masks and generate semantic segmentation results. Recent scribble-based WSSS domain approaches can be categorized into three main groups: regularization loss-based methods~\cite{tang2018normalized,tang2018regularized}, consistency learning-based methods~\cite{pan2021scribble,wang2022cycle}, and label diffusion-based methods ~\cite{lin2016scribblesup,vernaza2017learning,xu2021scribble}, as depicted in Figure~\ref{fig:ssss_intro}. Regularization loss-based methods aim to enhance network robustness by preventing it from being overconfident. Consistency learning-based methods leverage self-supervised learning strategies to acquire invariant features. While both these two kinds of methods contribute to enhanced network robustness, they still struggle to address the issue of lacking supervision. Specifically, BPG~\cite{wang2019boundary} utilizes extra boundary data with edge information to improve segmentation performance. Label diffusion-based approaches utilize scribbles to generate pseudo-labels using unsupervised models, such as graph models, and subsequently employ these pseudo-labels to train semantic segmentation models. However, such a diffusion process fails to effectively exploit the global semantic information lurking in the image. More recent works~\cite{liang2022tree,wu2023sparsely} aim to adaptively generate pseudo-labels using a tree filter and a learnable probabilistic model with Gaussian prior, respectively. Despite their advancements, both of these methods still lack image-level supervision, thereby limiting their ability to model global semantic information effectively.

\paragraph{Other WSSS Methods} Points~\cite{bearman2016s,chen2021seminar,wu2022deep,wu2023sparsely,liang2022tree} and bounding boxes~\cite{dai2015boxsup,papandreou2015weakly,khoreva2017simple,zhang2021affinity} are also common annotations in weakly-supervised semantic segmentation. However, both of them fail to achieve a balance between training supervision and labor costs. The point-level annotation requires less labor, but it provides very limited supervision, hence training a high-accuracy semantic segmentation model is difficult. Bounding boxes suffer from overlapping with each other when encountering multiple objects and provide redundant supervision, which may confuse the model. In comparison, scribbles achieve the best balance between laboring cost and supervision accuracy.

\section{Method}

In this part, we first retrospect the general problem formulation of label diffusion-based methods and their limitations. Then we introduce CDSP with the pseudo-label generation, basic supervision, LoRM, and DEL sequentially in detail.

\subsection{General Problem Formulation}

\label{sec:method}
Denoting $\mathbf{\Omega}=\{\boldsymbol{y}_i|i=1,...n\}$ as the ground truth label set and $\mathbf{\Omega}_{s}$ as the sparse scribble label, where $\mathbf{\Omega}_{s} \subset \mathbf{\Omega}$ and $|\mathbf{\Omega}_{s}| << |\mathbf{\Omega}|$. The objective function of the scribble-based WSSS can be formulated as:

\begin{equation}
  \min  c(\mathbf{P}_{\mathbf{\Omega}_s},\mathbf{\Omega}_s),
  \label{eq:ssss}
\end{equation}
where $c(\cdot, \cdot)$ denotes the criterion function, which is usually cross-entropy. $\mathbf{P}_{\mathbf{\Omega}_s}$ denotes the model predictions corresponding to the sparse scribble label.  Such sparse supervision limits the model's performance and decreases the certainty of the model. Most existing label diffusion-based methods make efforts on devising a graphical diffusion model or learnable probabilistic model with low-level cues $\phi$ to generate the pseudo-label $\widetilde{\mathbf{\Omega}}=\{\boldsymbol{\widetilde{y}}_i|i=1,...n\}$ by diffusing the labeled pixels to unlabeled ones:
\begin{equation}
  \widetilde{\mathbf{\Omega}} = \phi(\mathbf{\Omega}_{s}).
  \label{eq:diffuse}
\end{equation}
Combined with Eq~\ref{eq:ssss}, a complete objective function for scribble-based WSSS can be obtained:

\begin{equation}
  \min (c(\mathbf{P}_{\mathbf{\Omega}_s},\mathbf{\Omega}_s) +c(\mathbf{P}_{\mathbf{\widetilde{\Omega}}},\mathbf{\widetilde{\Omega}})).
  \label{eq:fsss}
\end{equation}
As shown in Eq.~\ref{eq:diffuse}, because only scribble-annotated pixels are considered, it is hard for the diffusion methods to capture the global information from the scribbles, making the diffused label $\boldsymbol{\widetilde{y}}$ provide locally considered supervision. Besides, it is evident that the diffused pseudo-label heavily depends on the scribble, where its quality may be undermined by a shrunk or dropped version of the scribble. 
\subsection{Class-driven Scribble Promotion}
To solve the problems mentioned above, we naturally think of utilizing the class label derived from the sparse scribble to provide global cues for image-supervised segmentation when generating the pseudo-label. Denoting $\widetilde{\phi}$ as the classification model with a fully connected layer, the pseudo-label $\widetilde{\mathbf{\Omega}}$ can be obtained from the image $\mathbf{I} \in \mathbb{R}^{3\times H \times W}$ with multi-class label $\boldsymbol{k}\in \mathbb{R}^{1\times K}$ :
\begin{equation}
    \boldsymbol{\widetilde{\Omega}} = \widetilde{\phi}(\mathbf{I},\boldsymbol{k}),
    \label{eq:pseudo}
\end{equation}
where all the pixels are taken into account to generate the pseudo-label. After that, we further introduce the LoRM and DEL to strike the advantages of both supervisions as shown in Figure~\ref{fig:shc}. In general, the overall loss function for supervision can be formulated as:
\begin{equation}
    \mathcal{L} = \mathcal{L}_{seg} + \mathcal{L}_{lorm} + \mathcal{L}_{de}.
    \label{eq:overall_loss}
\end{equation}
$\mathcal{L}_{seg}$ represents the basic supervision from the scribble and pseudo-label, $\mathcal{L}_{lorm}$ represents the supervision from the LoRM, and $\mathcal{L}_{de}$ is the supervision from DEL. The details of each component will be introduced sequentially in the following parts.

\subsection{Pseudo-label Generation and Basic Supervision}
To obtain the pseudo-label with Eq.~\ref{eq:pseudo}, we first train a multi-label classification model $\mathcal{C}(\cdot)$ followed by a $K$-class classifier (e.g. Resnet~\cite{he2016deep} with an FC layer) with image-level classes extracted from the scribbles. After the model converges, the image $\mathbf{I}$ is fed into the model to generate the class activate map of the $k^{th}$ class:
\begin{equation}
    CAM_{k} (\mathbf{I}) = ReLU( \sum_{i=1}^{C} \mathbf{W}_{i,k}\mathbf{F}_{i}),
\end{equation}
where $\mathbf{F}=\mathcal{C}(\mathbf{I}),\mathbf{F}\in \mathbb{R}^{C\times HW}$ is the feature maps of the last layer, $\mathbf{W}$ is the weight matrix in the classifier. We follow existing image-supervised semantic segmentation methods to threshold the $CAM$ into binary masks and integrate them into a single channel multi-class mask~\cite{wang2020self,chen2022class} to generate the pseudo-label $\widetilde{\mathbf{\Omega}}$. It is also possible to adopt one-stage image-supervised WSSS methods~\cite{ru2022learning,zhu2023branches} as $\widetilde{\phi}$ to generate the pseudo-label. With both pseudo-label and scribble, the basic supervision can be summarized as:
\begin{equation}
  \mathcal{L}_{seg} = \mathcal{L}_{segs} + \mathcal{L}_{segc}.
  \label{eq:l_seg}
\end{equation}
In detail, $\mathcal{L}_{segs}$ denotes the sparse supervision from the scribble label in the form of a partial cross-entropy:
\begin{equation}
    \mathcal{L}_{segs} = \frac{1}{|\mathbf{\Omega}_s|} \sum_{\boldsymbol{y}_i\in\mathbf{\Omega}_s} c(\boldsymbol{y}_i,\boldsymbol{p}_i),
\end{equation}
where $c(\boldsymbol{y}_i,\boldsymbol{p}_i)=-\sum_{k=1}^{K} \boldsymbol{y}_{i,k} log(\boldsymbol{p}_{i,k})$, $K$ is the class number, $\boldsymbol{p}_i$ is the prediction from the model, $\boldsymbol{y}_i$ is the one-hot label. $\mathcal{L}_{segc}$ denotes the supervision from the pseudo-label, which can be formulated as a smoothed cross-entropy:
\begin{equation}
    \mathcal{L}_{segc} = \frac{1}{|\mathbf{\widetilde{\Omega}}|} \sum_{\boldsymbol{y}_i\in \mathbf{\widetilde{\Omega}}}[(1-\epsilon)c(\boldsymbol{y}_i,\boldsymbol{p}_i)+\epsilon c(\frac{1}{K},\boldsymbol{p}_i)],
\end{equation}
where $\epsilon=0.1$ is a regularization item of label smoothing~\cite{muller2019does} to prevent the model from being over confident.
\subsection{Localization Rectification Module}
\begin{figure}[t]
  \centering
  \includegraphics[width=\linewidth]{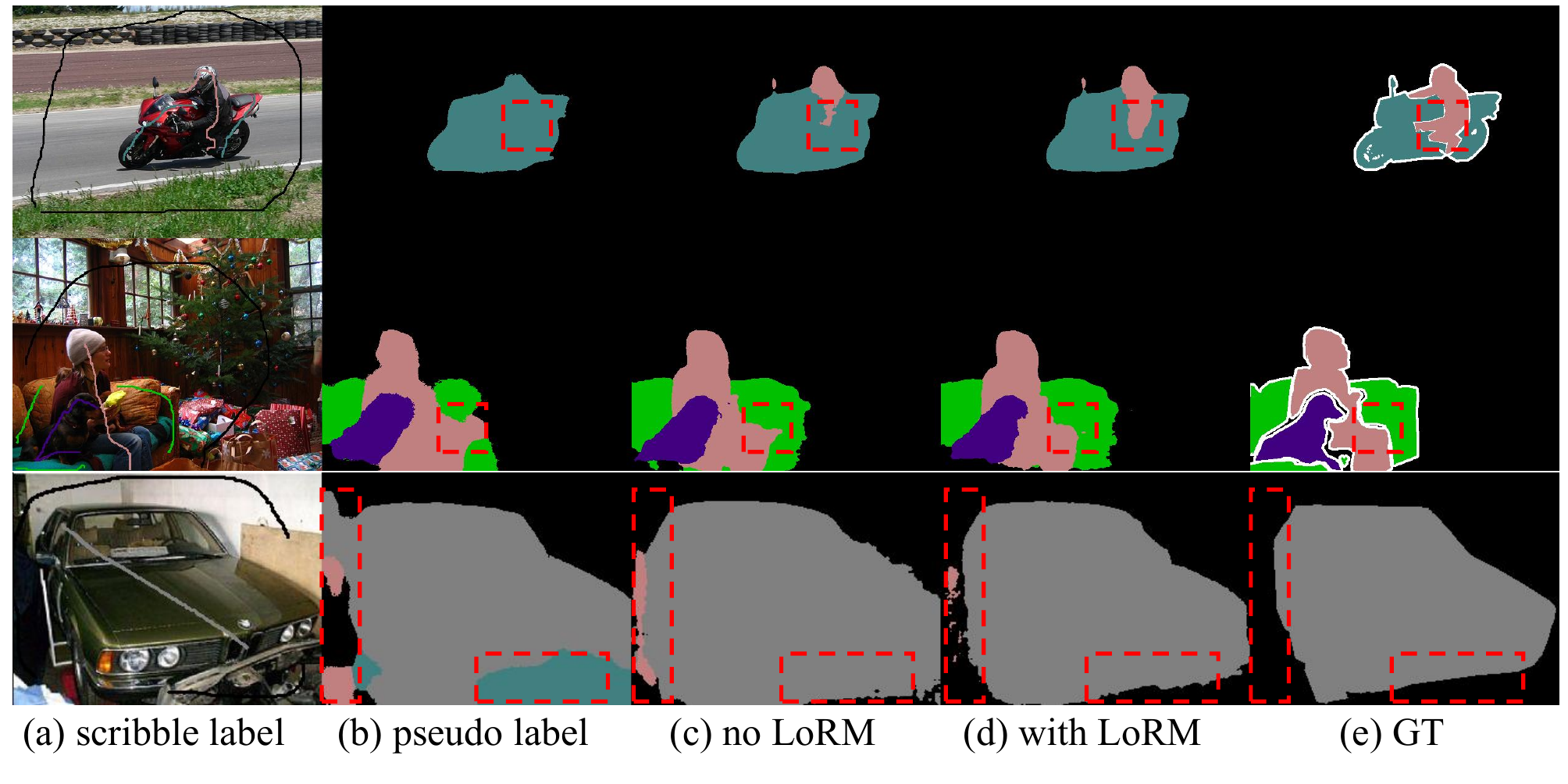}
  \caption{Visualization results employing resnet50 backbone and deeplabV2 segmentor. (a) is the original image with scribble label, (b) is the pseudo-label for training, (c) is the prediction trained with $\mathcal{L}_{seg}$, (d) is the prediction trained with $\mathcal{L}_{seg}+\mathcal{L}_{lorm}$. (e) is the ground truth label.}
  \label{fig:LoRM_vis}
\end{figure}

Adopting the pseudo-label directly for supervision can lead to absurd predictions~\cite{wang2018weakly}, particularly evident when foreground objects are nearby, as illustrated in Figure~\ref{fig:LoRM_vis}(c). Rather than correcting the pseudo-label itself, we are motivated to refine the feature representations of the model so that the model can adopt pseudo-labels with different qualities. To achieve this goal, we propose a novel module namely LoRM. The primary concept behind the LoRM is to leverage the inherent similarity of representations among foreground pixels belonging to the same semantic class. By doing so, mispredicted pixels can be refined through a weighted combination of representations from other pixels. Let $\mathbf{F} \in \mathbb{R}^{C\times H \times W}$ denotes the feature map generated by the last layer of the segmentation backbone $\mathcal{S}(\cdot)$, and $\mathbf{M} \in \mathbb{R}^{H\times W}$ denotes the pseudo mask, as depicted in Figure~\ref{fig:shc}. The LoRM takes $\mathbf{F}$ and $\mathbf{M}$ as inputs and operates accordingly to rectify the  representations.

\begin{figure}[h]
  \centering
  \includegraphics[width=0.9\linewidth]{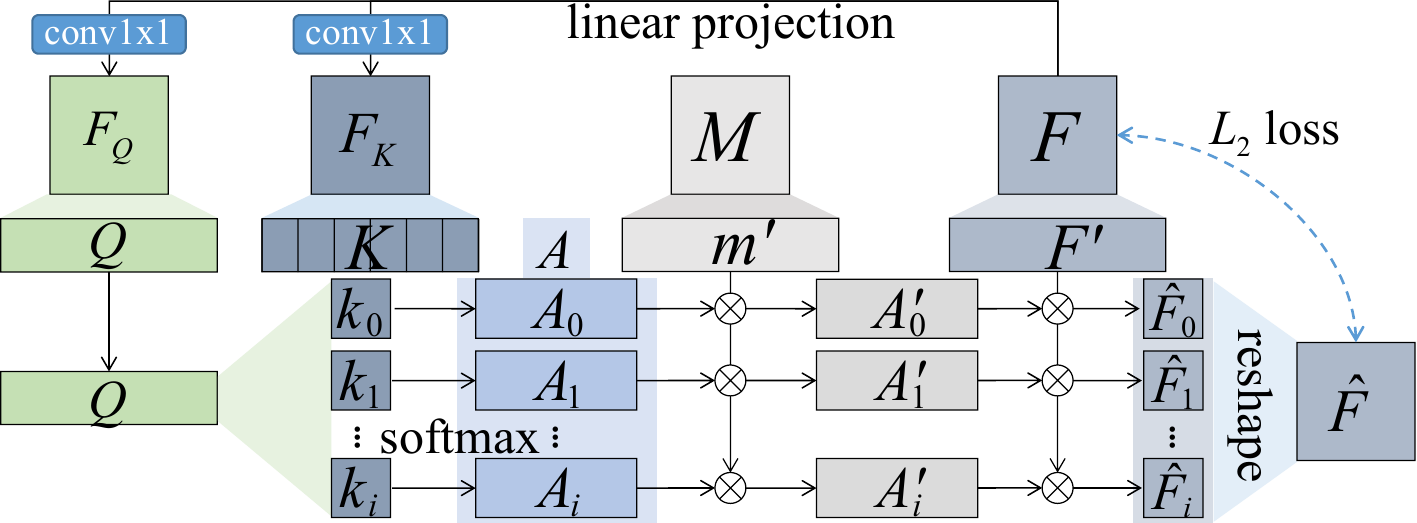}
  \caption{The illuastration of LoRM.}
  \label{fig:LoRM}
\end{figure}

As detailed in Figure~\ref{fig:LoRM}, the feature map $\mathbf{F}$ is firstly liner projected into $\mathbf{F}_Q\in \mathbb{R}^{C\times H \times W}$ and $\mathbf{F}_K\in \mathbb{R}^{C\times H \times W}$ with a single convolution, then flattened along the row axis into $\mathbf{Q}\in \mathbb{R}^{C\times HW}$ and $ \mathbf{K} \in \mathbb{R}^{C\times HW}$. Taking $\mathbf{K}$ as the key set to be refined, and $\mathbf{Q}$ as the query set for similarity matching, we calculate the weighted similarity matrix $\mathbf{A}$ by:
\begin{equation}
  \mathbf{A} = softmax(\frac{\mathbf{Q}^T \mathbf{K}}{\| \mathbf{Q}^T\Vert^{C}_2 \| \mathbf{K} \Vert^{C}_2}),
  \label{eq:similarity_A}
\end{equation}
where $\mathbf{A}\in \mathbb{R}^{HW\times HW}$, $softmax$ is implemented along the row axis, the L2-norm operation $\| \cdot \Vert_2^{C}$ of $\mathbf{Q}^T$ and $\mathbf{K}$ is implemented along the channel dimension. Each row $\mathbf{A}_i$ in the matrix $\mathbf{A}$ describes the similarity between the \emph{i-th} feature vector in $\mathbf{K}$ and all the $HW$ feature vectors in $\mathbf{Q}$. With the help of Eq.~\ref{eq:similarity_A}, the \emph{i-th} feature vector can be refined by referencing the feature vectors in other locations. It is worth noting that, the background vectors vary largely, and contribute little to the foreground rectification. Therefore, we extract the foreground mask $\mathbf{M}\in \mathbb{R}^{ H\times W}$ from the pseudo-label and flatten it along the row axis, then element-wise multiply it with $\mathbf{A}$ leveraging the broadcast technique:
\begin{equation}
  \mathbf{A'} = flatten(\mathbf{M}) * \mathbf{A},
\end{equation}
so that the background features in each row $\mathbf{A}_i$ are largely suppressed in its masked one $\mathbf{A'}_i$. Then the original feature map $\mathbf{F}$ is flattened along the row axis, and it is matrix-multiplied with the masked similarity matrix $\mathbf{A'}$:
\begin{equation}
  \mathbf{\hat{F}} = \delta * flatten(\mathbf{F}) \mathbf{A'} ,
\end{equation}
where $\delta$ is a learnable parameter initialized with 1 to control the rectification degree, $\mathbf{\hat{F}} \in \mathbb{R}^{C\times H W}$ is the refined feature which is finally reshaped back to $\mathbb{R}^{C\times H\times W}$. The mean square error loss (MSE) is implemented on the original feature $\mathbf{F}$ and the refined feature $\mathbf{\hat{F}}$:
\begin{equation}
  \mathcal{L}_{lorm} = MSE(\mathbf{F},\mathbf{\hat{F}}).
  \label{eq:l_LoRM}
\end{equation}
The whole process is realized by efficient matrix operations. With the supervision of Eq.~\ref{eq:l_LoRM}, the LoRM achieves the goal of rectifying the misled foreground representations by referencing the representations in other foreground locations.

\subsection{Distance Entropy Loss}
\label{sec:dem}

The LoRM effectively addresses the misalignment in the feature space in the foreground area, but the model remains susceptible to being misled by noisy labels near the object boundary during later training steps. This could undermine the efforts of LoRM and reduce the model's certainty. 

To overcome this challenge, it becomes crucial to identify reliable predictions. We propose that discriminative areas, such as the surroundings of the scribble, are more reliable and should be assigned higher confidence. Conversely, indiscriminative areas like the boundary of the pseudo-label, generated by global class supervision, are less reliable and should be assigned lower confidence. Based on this concept, we introduce a distance map strategy, to assign predictions with different confidence levels according to their distance from the scribble and the pseudo-label boundary respectively, known as the distance entropy loss. By doing so, we can better leverage the advantages of both supervisions during model training.

\begin{figure}[htbp]
  \centering
  \subfloat[Pseudo]{
      \includegraphics[width=0.22\linewidth]{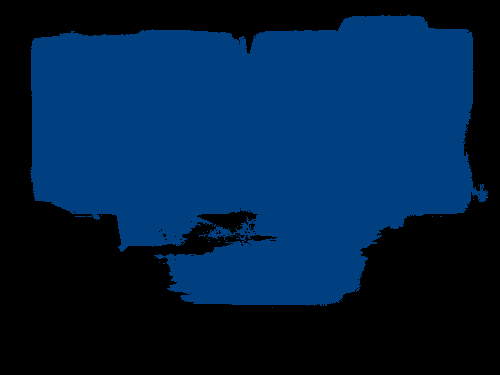}
      }
  \subfloat[$\lambda_c=1$]{
      \includegraphics[width=0.22\linewidth]{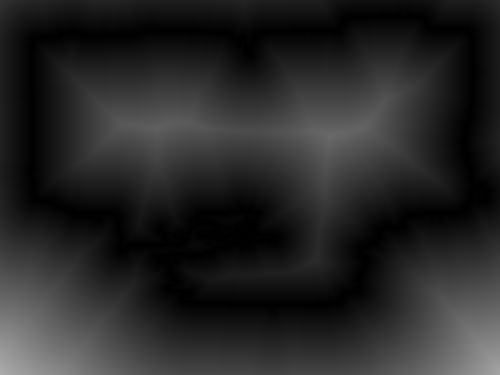}
      }
  \subfloat[$\lambda_c=e^3$]{
    \includegraphics[width=0.22\linewidth]{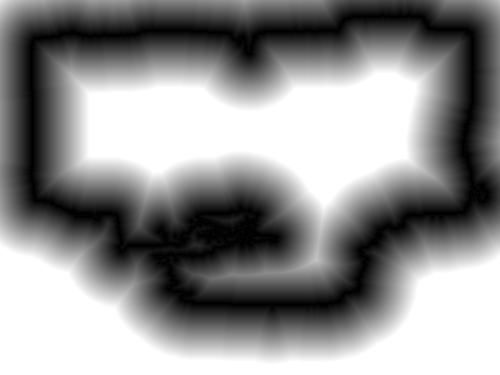}
      }
  \subfloat[$\lambda_c=e^7$]{
    \includegraphics[width=0.22\linewidth]{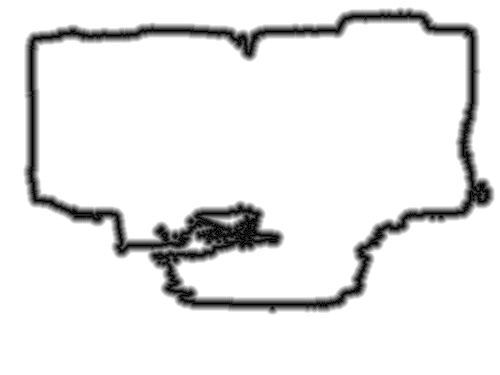}
      }
      
  \subfloat[Image]{
      \includegraphics[width=0.22\linewidth]{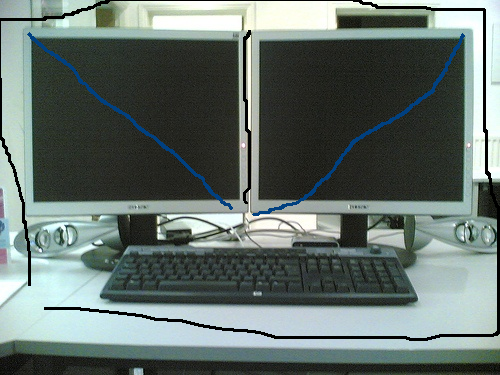}
      }
  \subfloat[$\lambda_s=1$]{
      \includegraphics[width=0.22\linewidth]{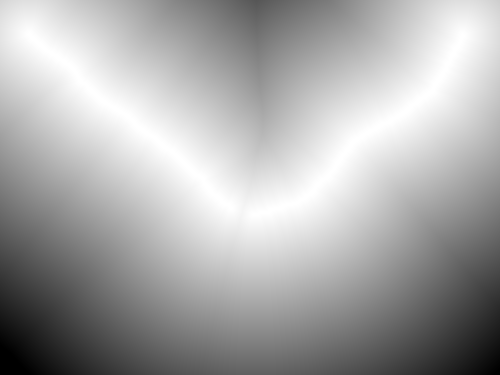}
      }
  \subfloat[$\lambda_s=e^3$]{
    \includegraphics[width=0.22\linewidth]{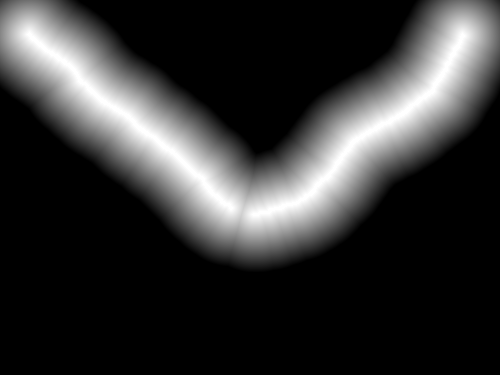}
      }
  \subfloat[$\lambda_s=e^7$]{
    \includegraphics[width=0.22\linewidth]{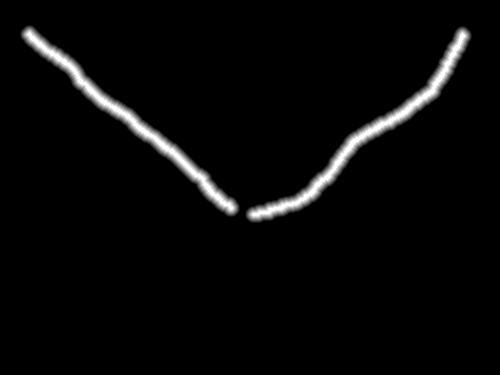}
      }

  \caption{ Visualization of disance maps with different coefficients for pseudo label boundary (b-d) and scribble (f-h)}
  \label{fig:DEM_vis}
\end{figure}

For the pseudo-label, the pixels around its boundary are indiscriminative, and such an area is probable to provide uncertain supervision. Denoting the coordinates of the $i^{th}$ point in the image as $(m,n)$, and the coordinates of the $j^{th}$ point on the foreground pseudo-label boundary as $(m',n')$, the distance maps of the pseudo-label is designed as:
\begin{equation}
  d_c(i)=\min\limits_{\forall j}(\frac{\lfloor \sqrt{e^{\lambda_c} [(m-m')^2+(n-n')^2]}\rfloor_{255}}{255}),
\end{equation}
where $d_c$ is a probability ranges in $[0,1]$ that describes the minimum Euclidean distance between a point and the set of pseudo-label boundary points with the distance value truncated to 255 for normalization and the efficiency of data storage. $\lambda_c$ is a coefficient to control the scope of the pseudo-label distance map as shown in Figure~\ref{fig:DEM_vis} (f-h). Denoting $N_c$ as the number of non-zero elements in $d_c$, the distance entropy of the pseudo-label is formulated as:
\begin{equation}
  \mathcal{L}_{dc} = \frac{1}{N_c} \sum_{i=1}^{N_c} d_c(i)\boldsymbol{p}_i log(\boldsymbol{p}_i).
  \label{eq:l_dc}
\end{equation}

Compared with the pseudo-label, the scribble is certain and correct, the pixels lying around the scribble may largely belong to the same semantic class as the scribble. Moreover, the scribble lying in the foreground's inner area provides correct supervision, which could suppress the noisy supervision in pseudo-label. But this confidence should decrease with the increment of the distance. Therefore, denoting the coordinates of the $i^{th}$ point in the image as $(m,n)$, and the $j^{th}$ foreground scribble point coordinates as $(m',n')$, the distance map of the scribble is designed as:
\begin{equation}
  d_s(i)=1-\min\limits_{\forall j}(\frac{\lfloor\sqrt{e^{\lambda_s} [(m-m')^2+(n-n')^2]}\rfloor_{255}}{255}),
\end{equation}
where $d_s$ is a probability ranges in $[0,1]$ that describes the minimum Euclidean distance between a point and the set of scribble points. $\lambda_s$ is a coefficient to control the scope of the scribble distance map as shown in Figure~\ref{fig:DEM_vis}(b-d). Denoting $N_s$ is the number of nonzero elements in $d_s$, the distance entropy of the scribble is formulated as:
\begin{equation}
  \mathcal{L}_{ds} = \frac{1}{N_s} \sum_{i=1}^{N_s} d_s(i) \boldsymbol{p}_i log(\boldsymbol{p}_i),
  \label{eq:l_ds}
\end{equation}
Finally, the overall distance entropy can be formulated as:
\begin{equation}
  \mathcal{L}_{de} = \mathcal{L}_{ds}+\mathcal{L}_{dc}.
  \label{eq:l_dem}
\end{equation}
Figure~\ref{fig:DEM_vis} presents visualizations of the distance maps for the scribble and pseudo-label boundaries at different coefficients of $\lambda_s$ and $\lambda_c$. As $\lambda_s$ increases, the reliable area determined by the scribble becomes more prominent. Conversely, a higher $\lambda_c$ endows more weights to the pseudo-label in determining the reliable area. Through the distance entropy loss, we effectively excavate the reliable areas and reinforce the prediction certainty of the model by leveraging information from both the scribble and the pseudo-label boundaries.

\begin{table}[ht]
\centering
\tabcolsep=0.05cm
\begin{tabular}{lclcc} 
\toprule
Method & Sup & Segmentor & val & test  \\ 
\hline
AFA~\cite{zhang2021affinity}& $\mathcal{I}$  & SegFormer & 66.0  &- \\
AMN~\cite{lee2022threshold} & $\mathcal{I}$  & r101+v2 & 70.7  &- \\
BECO~\cite{rong2023boundary}& $\mathcal{I}$  & MiT+v3p & 73.7  &- \\
TOKO~\cite{ru2023token}     & $\mathcal{I}$  & ViT+v2  & 72.3  &- \\
\hline
BoxSup (Dai et al. 2015) & $\mathcal{B}$  & vgg16+v1      & 62.0 &- \\
WSSL~\cite{papandreou2015weakly} & $\mathcal{B}$ & vgg16+v1  & 67.6 &-  \\
SDI~\cite{khoreva2017simple} & $\mathcal{B}$ & vgg16+v1      & 65.7 &- \\
BBAM~\cite{lee2021bbam}     & $\mathcal{B}$  & r101+v2       &  63.7  &-\\
\hline
ScribbleSup (Lin et al. 2016) & $\mathcal{S}$   & vgg16+v1   & 63.1    &-  \\
RAWKS (Vernaza et al. 2017) & $\mathcal{S}$   & r101+v1  & 61.4  &- \\
NCL (Tang et al. 2018a) & $\mathcal{S}$   & r101+v1  & 72.8   &- \\
KCL (Tang et al. 2018b) & $\mathcal{S}$   & r101+v2  & 72.9  &- \\
BPG (Wang et al. 2019) & $\mathcal{S}$    & r101+v2  & 73.2  &- \\
PSI (Xu et al. 2021) & $\mathcal{S}$     & r101+v3p & 74.9   &- \\
URSS (Pan et al. 2021) & $\mathcal{S}$      & r101+v2  & 74.6 & 73.3 \\
CCL (Wang et al. 2022)  & $\mathcal{S}$  & r101+v2  & 74.4   &- \\
TEL (Liang et al. 2022) & $\mathcal{S}$     & r101+v3p & 75.2  & 75.6   \\
AGMM (Wu et al. 2023)  & $\mathcal{S}$     & r101+v3p & 74.2  & 75.7 \\ 
\hline
Ours        & $\mathcal{S}$          & r50+v2 & 73.9   & 74.2   \\
Ours        & $\mathcal{S}$          & r101+v2 & 75.3  & 75.3   \\
Ours        & $\mathcal{S}$          & r101+v3p & \textbf{75.9}  & \textbf{76.0}    \\
\multicolumn{1}{l}{baseline (scribble only)} & $\mathcal{S}$  & r101+v3p & 66.2  & 69.7  \\ 
\bottomrule
\end{tabular}
\caption{Comparison with the state-of-the-arts methods.}
\label{tab:sota}
\label{tab:wsss_voc}
\end{table}

\begin{figure*}[ht]
    \centering
    \subfloat[I+S]{\includegraphics[width=0.12\linewidth]{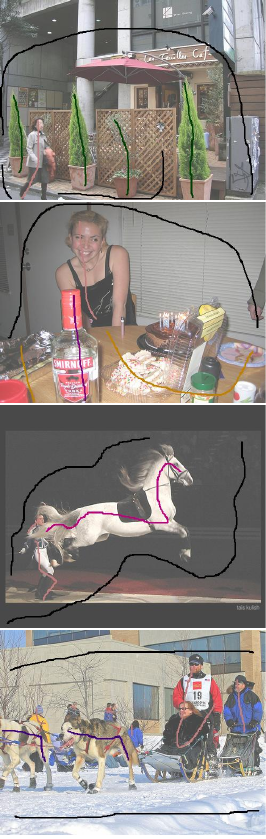}}
    \subfloat[baseline]{\includegraphics[width=0.12\linewidth]{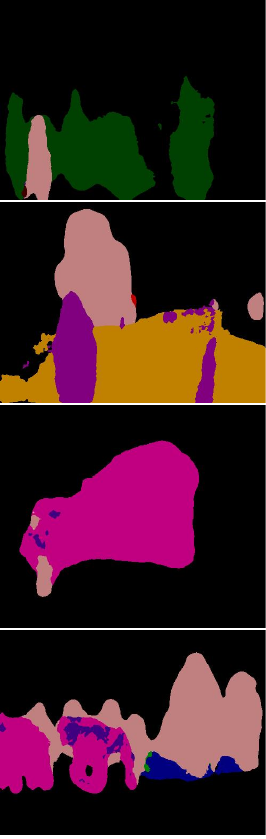}}
    \subfloat[URSS]{\includegraphics[width=0.12\linewidth]{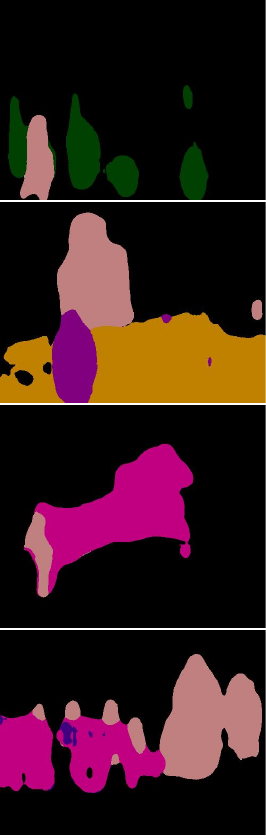}}
    \subfloat[TEL]{\includegraphics[width=0.12\linewidth]{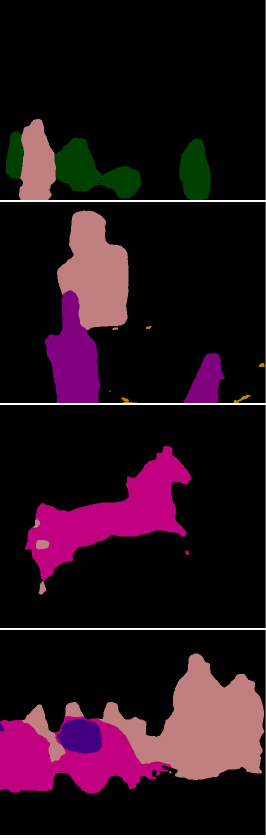}}
    \subfloat[AGMM]{\includegraphics[width=0.12\linewidth]{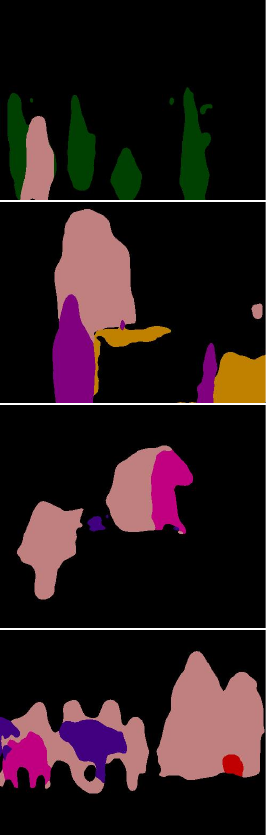}}
    \subfloat[Ours]{\includegraphics[width=0.12\linewidth]{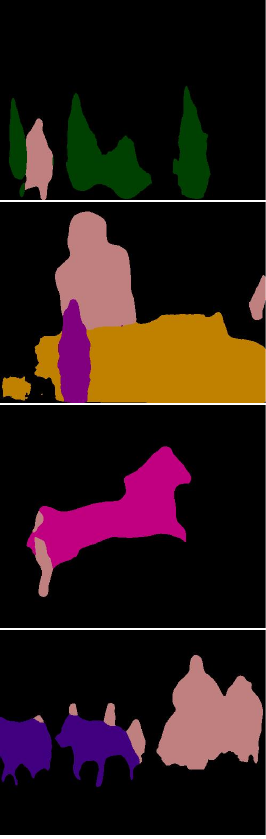}}
    \subfloat[GT]{\includegraphics[width=0.12\linewidth]{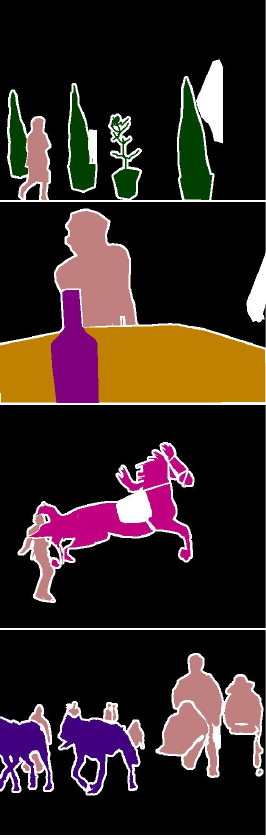}}
    \caption{Visualization results comparison. (a) is the image with its scribble annotations. The baseline (b) is deeplabV3+ trained with only scribble annotations. (c) to (e) are recent methods, and (g) is the ground truth label.}
    \label{fig:compare_visualization}
\end{figure*}

\section{Experiments}
\paragraph{Dataset}
Our experiments were carried out on the widely used ScribbleSup dataset~\cite{lin2016scribblesup}, which combines PASCAL VOC2012 and SBD ~\cite{hariharan2011semantic} datasets with scribble annotations. The dataset includes 10,582 training images and 1,449 validation images. To ensure fairness, we used the same scribble generation code as previous works~\cite{lin2016scribblesup,tang2018regularized,pan2021scribble}, maintaining uniform scribble thickness. Additionally, we validated our method on \emph{scribble-shrink} and \emph{scribble-drop} introduced by URSS~\cite{pan2021scribble} to assess its robustness in diverse scenarios.
\paragraph{Implementation Details}

With the pseudo-labels generated by BMP~\cite{zhu2023branches}, we employed representative segmentation frameworks deeplabV2~\cite{chen2017deeplab} and deeplabV3+~\cite{chen2018encoder} for method validation and generating competitive results, respectively. We conducted a total of 50 epochs with a base learning rate of $1e^{-3}$ and batch size set to 16 for training. To ensure stable training, we adopted a learning rate warmup strategy, linearly increasing the learning rate to $1e^{-3}$ over the first 10 epochs, followed by a cosine decay to zero over the next 40 epochs. Validation results were reported using the last checkpoint. The stochastic gradient descent (SGD) optimizer was utilized with a momentum of 0.9 and weight decay of $5e^{-4}$. Data augmentation followed the same strategy used in URSS. All experiments were reported with the mIoU metric (\%) and conducted on one NVIDIA RTX 4090 24G GPU with an Intel Xeon Gold 6330 CPU. 

\paragraph{Comparison on ScribbleSup}

\begin{table}[ht]
    \centering
    \begin{tabular}{cccccc} 
    \toprule
    \multicolumn{2}{c|}{basic supervision}  & \multicolumn{2}{c|}{$\mathcal{L}_{de}$}          & \multirow{2}{*}{$\mathcal{L}_{lorm}$}  & \multirow{2}{*}{mIoU}  \\ 
    \cline{1-4}
    $\mathcal{L}_{segs}$  & \multicolumn{1}{c|}{$\mathcal{L}_{segc}$}  & $\mathcal{L}_{ds}$  & \multicolumn{1}{c|}{$\mathcal{L}_{dc}$}  & & \\ 
    \hline
    \checkmark & \multicolumn{4}{c}{}  & 66.17                       \\
             & \checkmark & & & & 67.23                       \\
    \checkmark & \checkmark & & & & 72.13                       \\
    \hline
    \checkmark & \checkmark & & \checkmark & & 67.33                       \\ 
    \checkmark & \checkmark & \checkmark & & & 73.38                       \\
    \checkmark & \checkmark & \checkmark & \checkmark & & 73.58                       \\
    \hline
    \checkmark & \checkmark & & & \checkmark &  73.26 \\ 
    \checkmark & \checkmark & \checkmark & & \checkmark & 73.51 \\
    \checkmark & \checkmark & & \checkmark & \checkmark & 73.64 \\
    \checkmark & \checkmark & \checkmark & \checkmark & \checkmark & \textbf{73.91}              \\
    
    \bottomrule
    \end{tabular}
    \caption{The effectiveness of each component.}
    
    \label{tab:ablation}
    \end{table}
\begin{figure*}[htbp]
  \centering
  \subfloat[scribble-drop]{
      \includegraphics[width=0.45\linewidth]{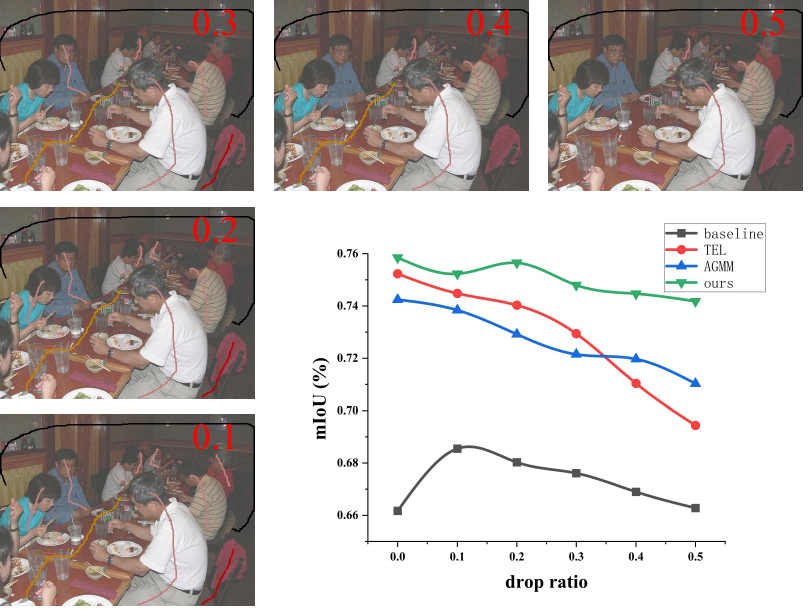}
      }
  \subfloat[scribble-shrink]{
      \includegraphics[width=0.45\linewidth]{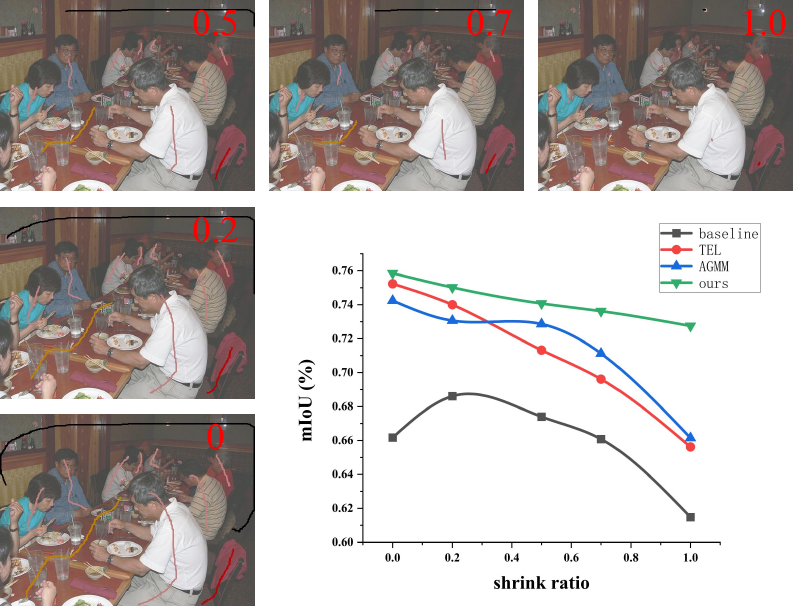}
      }
  \caption{The experiments on scribble-drop and scribble-shrink dataset with different drop or shrink ratios.}
  \label{fig:shrink_drop}
\end{figure*}
We deploy resnet101~\cite{he2016deep} as the backbone and deeplabV3+ as the segmentor with hyper-parameters of $(\lambda_s=e^2,\lambda_c=e^7)$ to generate the best result. The comparison details are recorded in Table~\ref{tab:sota}. It is worth noting that, previous works of ScribbleSup, RAWKS~\cite{vernaza2017learning}, and NCL~\cite{tang2018normalized} adopted CRF for postprocessing, which was fairly time-consuming. For recent works of TEL~\cite{liang2022tree} and AGMM~\cite{wu2023sparsely}, they were designed for general sparsely supervised segmentation, covering point level, scribble level, and box level annotations. To ensure the fairness, we reimplemented them using standard scribbles commonly used in previous works like ScirbbleSup, NCL, and URSS. As shown in Table~\ref{tab:sota}, our method outperforms all the previous methods, exceeding the TEL by 0.6\% and AGMM by 1.6\%. The test results reported in the last column of Table~\ref{tab:sota} are acquired from PASCAL VOC2012 website~\cite{everingham2012pascal}. The visualization comparison of our method using deeplabV3+ with previous SOTA methods is shown in Figure~\ref{fig:compare_visualization}, where recent methods fail to capture correct global semantics.
\paragraph{Shrink and Drop}
As scribble-based annotations are flexible, it is common that the user annotates the scribbles with different length and sometimes drop some of the objects. Therefore, evaluating the model's robustness with different shrink or drop ratios is also essential. Some shrunk or dropped samples are presented in Figure~\ref{fig:shrink_drop}. Notably, as depicted in the figure, an increase in the drop or shrink ratio leads to a decrease in the model's performance. Specifically, when the scribbles are shrunk to points ($shrink\ ratio=1$), AGMM and TEL experience an approximately 10\% performance degradation. In contrast, our method exhibits only a marginal drop within 1\%, showcasing its robustness. 

\paragraph{Ablation on Components} We employ resnet50 backbone with deeplabV2 as the segmentor and use the ScribbleSup~\cite{lin2016scribblesup} dataset for training and validation. The optimal hyper-parameter combination of the distance entropy loss with all components is found by grid-search, where $\lambda_s=1,\lambda_c=6$, then we validate the effectiveness of each module by eliminating them one by one. The results are recorded in Table~\ref{tab:ablation}. It can be observed from the first three lines that, employing either scribble or pseudo-label as the basic supervision generates an unsatisfied result (only around 67\%), while using both of them produces a much better result (72.13\%). This demonstrates that the scribble and pseudo-label provide complementary supervision and they compensate each other. Additionally, only adding $\mathcal{L}_{dc}$ on the basic supervision degrades the model to almost the same performance as merely using $\mathcal{L}_{segc}$. This issue is attributed to the overfitting of the noisy labels in pseudo-labels of the model and can be addressed by our LoRM, which improves the model performance from 67.33\% to 73.64\%. Compared with the baseline, all the components obtain a better performance, and using them all achieves the best performance.
\paragraph{Ablation on Pseudo-labels}We also conducted experiments with different pseudo-labels to assess their influence, utilizing deeplabV3+ as the segmentor. The results in Table~\ref{tab:different_cams} indicate that, as the pseudo-label base accuracy improves, our method exhibits increasing performance. This demonstrates that our approach directly benefits from image-level WSSS methods, making it a promising avenue for further development.



\begin{table}[h]
    \centering
        \begin{tabular}{llcc} 
        \toprule
        Method      & Base acc & \multicolumn{1}{l}{res50} & \multicolumn{1}{l}{res101}  \\ 
        \hline
        SEAM (Wang et al. 2020)  & 64.5 & 69.8                  & 71.8                    \\
        AFA  (Ru et al. 2022)  & 66.0 & 71.5                  & 73.3                    \\
        BMP  \citep{zhu2023branches} & 68.1 & 73.9                  & 75.9                    \\
        \bottomrule
        \end{tabular}
    \caption{Performance adopting different pseudo-labels.}
    
    \label{tab:different_cams}
\end{table}
\section{Conclusion}
We propose a class-driven scribble promotion network for the scribble-based WSSS problem. To address the issue of model overfitting to noisy labels, we introduce a localization rectification module. Additionally, a distance entropy loss is incorporated to enhance the robustness of the network. Experimental results show that our method outperforms existing approaches, achieving state-of-the-art performance. 
\section{Acknowledgements}
This work was supported in part by the Natural Science Foundation of China (82371112, 62394311, 62394310), in part by Beijing Natural Science Foundation (Z210008), and in part by Shenzhen Science and Technology Program, China (KQTD20180412181221912).


\bibliography{aaai24}

\end{document}